\documentclass{svproc}

\usepackage{url}

\usepackage{graphicx}
\usepackage{float}
\usepackage{amsmath}
\usepackage{amssymb}   
\usepackage{booktabs}
\usepackage{tikz}
\usetikzlibrary{arrows.meta,positioning}
\graphicspath{{figures/}{./}}

\begin{document}
\mainmatter

\title{Asleep at the Wheel: JEPA's Limitations in Evaluating Novel Driving Data}
\titlerunning{Asleep at the Wheel}

\author{Advait Pavuluri$^{\spadesuit\blacklozenge\clubsuit}$,
Shamik Karkhanis$^{\spadesuit\blacklozenge\clubsuit}$,
Uzma Mushtaque$^{\clubsuit}$}
\authorrunning{Advait Pavuluri, Shamik Karkhanis, Uzma Mushtaque}
\institute{Rensselaer Polytechnic Institute, Troy, NY, USA}

\maketitle

\let\thefootnote\relax\footnotetext{$\spadesuit$\,Equal contribution. Listing order is completely random and both authors share first authorship.}
\let\thefootnote\relax\footnotetext{$\blacklozenge$\,Work done through the Undergraduate Research Program at Rensselaer Polytechnic Institute.}
\let\thefootnote\relax\footnotetext{$\clubsuit$\,Corresponding: \texttt{\{pavula, mushtu\}@rpi.edu}, \texttt{shamikkark@gmail.com}}

\begin{abstract}
Modern autonomous-driving fleets record far more video than human reviewers can inspect. This motivates the need for an automatic clip triage mechanism, to surface rare and review-worthy clips, so that driving models can be fine-tuned to better handle unideal circumstances. We test a label-free approach that scores clips by the prediction-error ``novelty'' of a self-supervised joint-embedding predictive architecture (JEPA); a frozen V-JEPA video encoder is paired with a lightweight predictor head to reconstruct masked clip embeddings, and clips whose embeddings are hard to predict are flagged as interesting. Evaluated under a realistic protocol that trains on one dataset and tests against footage from others, this approach appears highly effective. We show that this apparent success is actually a domain-shift consequence: on a fair benchmark drawn from a single dataset, this mechanism collapses to chance and is on par with simple no-training baselines. A lightly supervised probe on the same frozen embeddings results in almost double the average precision, indicating that the bottleneck is indeed the self-supervised objective, rather than the representation. We present this as a study for evaluating the effectiveness of self-supervised learning, where cross-dataset protocols can silently reward domain separation over novelty.
\end{abstract}

\section{Introduction}

As autonomous-driving fleets, such as Tesla and Waymo vehicles, have risen in popularity, many have become increasingly concerned about the data used to train their underlying machine learning models. Though these vehicles have gotten better at navigating in ideal conditions, they (similar to humans) struggle when unexpected circumstances arise, such as when encountering jaywalking or construction sites. As a result, experts have become interested in fine-tuning these driving models by using footage that contains such events.

However, modern autonomous-driving programs record orders of magnitude more video than
any team can inspect. A single fleet vehicle can produce dozens of hours of
camera footage per week, the overwhelming majority of which depicts routine,
uneventful driving. The clips that actually warrant human attention (reckless driving, unexpected maneuvers, breaking the rules of the road, etc.) are buried in this stream. Therefore, \emph{clip triage}, the process of sorting through hours of this footage and ranking unlabeled clips so that only the review-worthy ones rise to the top, is the problem we are trying to solve without human intervention. Doing so autonomously greatly reduces the human effort needed to curate training and evaluation data.

A natural way to attack triage without labels is to treat it as
a \emph{self-supervised} novelty estimation problem. Joint-embedding predictive architectures
(JEPAs) learn to predict the representation of a masked part of an input from
the visible part, and the resulting prediction error is an intuitive proxy for
``interestingness.'' In other words, content the model has not internalized should theoretically be harder to predict.

Building on Meta's frozen V-JEPA~2 video encoder~\cite{vjepa2}, we attach a lightweight
predictor head and train it with a masking objective, then score each clip by
the error between the predicted and the true embedding of the masked region. We
call the resulting quantity a \emph{review-value score}, the hypothesis being
that high prediction error flags clips a reviewer should see.

Evaluated in the way such systems are most commonly demonstrated, this approach
looks excellent. When the positive (review-worthy) clips are drawn from one set
of driving datasets and the negatives from another, the trained predictor
separates them with Average Precision (AP) near $0.89$ against a chance level of
$0.50$~\cite{nuscenes,waymo,bdd100k}. Taken at face value, this would be a
strong unsupervised triage result.

It is not. The positives and negatives in this setup differ
not only in whether they are review-worthy, but also in \emph{which dataset they
came from}. We show that the trained
predictor's novelty score is, to a very good approximation, a dataset
classifier: it separates Waymo/BDD100K clips from nuScenes clips with an area
under the ROC curve (AUC) of $0.965$, almost as well as a logistic classifier
trained directly on the frozen embeddings to predict provenance (AUC~$1.00$).
The high cross-dataset AP measures domain shift, not review value.

When we remove the confound and re-evaluate the \emph{same checkpoint} on a fair
benchmark whose positives and negatives are both drawn from nuScenes, the
apparent advantage collapses: Average Precision falls to $0.29$ against a chance
floor of $0.24$. Larger pretext-training corpora,
denser temporal sampling, and several no-training baselines (a masked-gap score,
a kNN-density score, and an untrained predictor) all behave the same way---none
meaningfully beats the positive rate. The unsupervised masking objective, as
instantiated here, does not surface the triage signal.

Crucially, a \emph{supervised} logistic probe trained on the same embeddings
reaches AP $0.50$ on the fair benchmark, roughly double the chance
floor. Therefore, the frozen encoder \emph{does} encode information relevant to review
value; the masking-based novelty objective simply fails to extract it, and
instead latches onto whatever varies most across the data---here, dataset
provenance. We therefore present this work as a diagnostic
study rather than as a new method that would work in a real-world scenario. While self-supervised learning seems promising at first glance, in this scenario it lacks nuance that fully-supervised learning is able to capture. Our evidence comes from one concrete configuration---a frozen V-JEPA~2 encoder with a lightweight masked-embedding predictor, a single masking strategy, and a single evaluation protocol---so we read the result as a property of this setup and its evaluation rather than a blanket verdict on masking-based self-supervision, and we return in Section~\ref{sec:limitations} to how far it might generalize.

\smallskip
\smallskip
\noindent\textbf{Contributions.}
\begin{itemize}
  \item We document a concrete and easily reproduced \emph{evaluation
        confound}: a frozen V-JEPA~2 predictor scores AP~$0.89$ on a
        cross-dataset triage benchmark but AP~$0.29$ (near chance) on a fair
        within-nuScenes benchmark, using the identical trained checkpoint.
  \item We show that the trained-predictor novelty score functions as a
        \emph{domain detector}, separating Waymo/BDD from nuScenes at AUC
        $0.965$ versus AUC $1.00$ for a direct embedding-based domain
        classifier, explaining the inflated cross-dataset number.
  \item We provide a \emph{supervised diagnostic} showing the triage signal is
        nonetheless present in the frozen embeddings (AP~$0.50$ vs.\ a $0.24$
        floor), localizing the failure to the masking objective rather than to
        the representation.
  \item We provide an \emph{open-source framework} to run these experiments, fully streamlining clip ingestion and allowing a user to run any number of end-to-end JEPA-style experiments.
\end{itemize}

\smallskip
\noindent The remainder of the paper is organized as follows.
Section~\ref{sec:related} situates the
work within masked autoencoding, joint-embedding predictive architectures, and
video anomaly detection, and states the gap we address.
Section~\ref{sec:method} formalizes the triage problem and describes the frozen
encoder, the masked-embedding predictor, the novelty score, and the supervised
and no-training baselines. Section~\ref{sec:setup} details the datasets,
benchmarks, labeling, metrics, and compute. Section~\ref{sec:results} presents
the cross-dataset versus fair-benchmark contrast, the domain-detector analysis,
and the configuration sweep. Sections~\ref{sec:discussion}
and~\ref{sec:limitations} discuss why the objective fails, the broader
implications for evaluating self-supervised triage, and the limitations that
scope our conclusions, before Section~\ref{sec:conclusion} concludes.

\section{Related Work}
\label{sec:related}

Having framed clip triage as label-free novelty estimation, we now place our
approach among the lines of work it draws on. Our method inherits the masking
objective of masked autoencoders, moves the prediction target into
representation space as joint-embedding predictive architectures do, and, at
inference, produces what is effectively a video anomaly score. We review these
three threads in turn.

\subsection{Masked Autoencoders}
\begin{figure}[tbp]
\centering
\includegraphics[width=\textwidth]{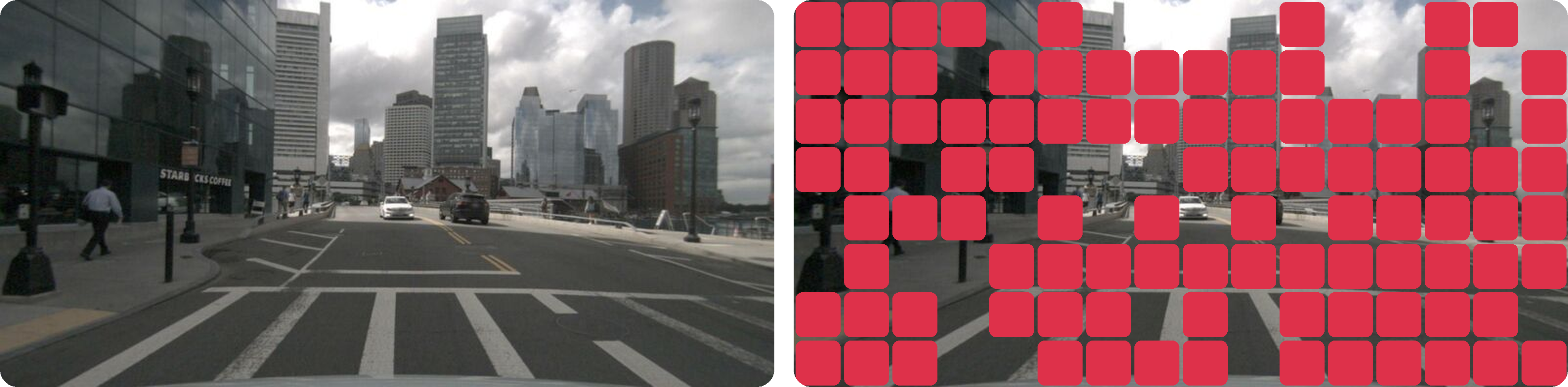}
\caption{Masked autoencoding on a driving frame. A large fraction of patches
(here $75\%$, red) is hidden (left: original; right: masked view) and the model
predicts what was removed. MAE~\cite{mae} and VideoMAE~\cite{videomae}
reconstruct pixels; the JEPA we build on predicts in latent space.}
\label{fig:mae}
\end{figure}

\textit{Masked Autoencoders Are Scalable Vision Learners}~\cite{mae} is a
2021 landmark paper in computer vision, later published at CVPR~2022. The paper introduced a simple
self-supervised learning approach: mask a large random fraction of the patches in an image
and train the model to reconstruct the missing pixels from the few that remain
(Figure~\ref{fig:mae}). Two design choices make this scale. The first is an
asymmetric encoder--decoder. The encoder is a Vision Transformer that processes
only the visible patches, while a lightweight decoder rebuilds the full image
from the encoded patches together with learned mask tokens. Because the encoder
never sees mask tokens, it runs on a small fraction of the input, which lowers
pretraining cost by roughly $3\times$ and makes very large models practical. The
second is an aggressive masking ratio: hiding about $75\%$ of the patches removes
enough redundancy that reconstruction stops being local interpolation and starts
to require a more holistic understanding of the image. Pretrained this way on
ImageNet-1K, a plain ViT-Huge reaches $87.8\%$ top-1 accuracy after fine-tuning
and transfers better than supervised pretraining to detection and segmentation.
VideoMAE~\cite{videomae} carries the recipe to clips with even higher space--time
masking ratios, and multimodal variants such as M3AE~\cite{m3ae} apply the same
reconstruct-the-missing objective across images and text. These methods rebuild
the raw input in pixel space; MaskFeat~\cite{maskfeat} takes a step away from
pixels by predicting feature descriptors of the masked spatiotemporal regions
instead, moving the prediction target from pixels toward representations.

\subsection{Joint-Embedding Predictive Architectures}
Joint-embedding predictive architectures (JEPAs) keep the masking objective, but move
the prediction target out of pixel space and into representation space. Instead
of reconstructing the missing pixels, a JEPA predicts the \emph{latent
representation} that an encoder would assign to the masked region.
I-JEPA~\cite{ijepa}, the image instantiation, makes this concrete with three
parts: a context encoder $f_\theta$ that embeds the visible block into
representations $\boldsymbol{s}_x$; a target encoder $f_{\bar\theta}$ that embeds the full
image into patch-level targets $\boldsymbol{s}_y=\{\boldsymbol{s}_{y_1},\dots,\boldsymbol{s}_{y_N}\}$; and a predictor
$g_\phi$ that, given $\boldsymbol{s}_x$ and a mask token $m_j$ for each masked patch, outputs a
prediction $\hat{\boldsymbol{s}}_y(i)$ for target block $i$. Training minimizes the average
$L_2$ distance between predicted and target representations,
\begin{equation}
  \frac{1}{M}\sum_{i=1}^{M} D\!\left(\hat{\boldsymbol{s}}_y(i),\, \boldsymbol{s}_y(i)\right)
  \;=\; \frac{1}{M}\sum_{i=1}^{M}\sum_{j\in B_i}
        \bigl\|\hat{\boldsymbol{s}}_{y_j} - \boldsymbol{s}_{y_j}\bigr\|_2^2,
  \label{eq:jepa}
\end{equation}
over the $M$ target blocks, with $B_i$ the set of patches in block $i$. The
targets come from a stop-gradient branch: $f_{\bar\theta}$ is an exponential
moving average of $f_\theta$, its momentum rising from $0.996$ toward $1$, which
keeps the latent objective from collapsing to a constant. Predicting in
representation space lets the model discard unpredictable pixel detail and
concentrate on more abstract, semantic structure, an aim contrastive
methods~\cite{contrastive_video_ssl} pursue from a different direction.
Extending these ideas to video also means modeling temporal structure;
TimeSformer~\cite{timesformer} showed that transformer self-attention can be
applied jointly across space and time for video understanding. V-JEPA~\cite{vjepa}
brought the joint-embedding predictive objective to video, predicting masked
spatiotemporal regions in a learned representation space, and
V-JEPA~2~\cite{vjepa2} scaled it up to support video understanding, prediction,
and planning. We build on the frozen V-JEPA~2 encoder, and the prediction residual
it produces is precisely the quantity we test as a review-value score in the rest
of the paper. Where prior JEPA work evaluates the learned representation through
downstream recognition, anticipation, or planning benchmarks, we instead ask
whether the prediction residual itself ranks driving clips for human review.

\subsection{Video Anomaly Detection}
Our novelty score is essentially an anomaly score: it flags the clips the model
predicts worst. This places our work squarely within video anomaly detection
(VAD), where a model fit to normal footage flags what it cannot explain; the
survey of Bogdoll et al.~\cite{bogdoll_ad_survey} catalogues this large body of
work for autonomous driving across sensing modalities and supervision regimes.
The supervision available varies widely: Sultani et al.~\cite{sultani_anomaly}
learn from videos labeled only as normal or anomalous, driving-specific datasets
such as DoTA~\cite{dota} provide spatial, temporal, and categorical anomaly
annotations, and recent pipelines push toward fine-grained, object-level
detection~\cite{tao_vad}. Self-supervised formulations close to ours also
exist---Orlova et al.~\cite{ssl_tad}, for instance, build traffic anomaly
detectors on frozen video foundation models with a masked-video-modeling
objective---so we do not claim the absence of labels as a distinguishing
feature. Our
clip-level scores are themselves a JEPA-residual instance of the familiar
``model the normal, flag the unexplained'' recipe, and our evaluation still draws
on labeled driving datasets. Where most of this work targets specific anomalous
objects or events, our aim is to rank whole clips for review. What has gone
unexamined, though, is a failure mode of the evaluation itself: across these
threads a score is typically judged on data whose ``normal'' and ``anomalous''
partitions come from the same source, or gains are reported without isolating
what the score actually keys on. A prediction-error novelty score evaluated under
the intuitive train-on-one-dataset, test-against-others protocol can instead
score well by detecting \emph{provenance} rather than review value, and to our
knowledge no prior study separates that confound from a same-source measurement.
Rather than proposing a new detector, our contribution is to expose this confound
and to show that the underlying triage signal is nonetheless recoverable from the
frozen representation.
\section{Method}
\label{sec:method}

\begin{figure}[tbp]
\centering
\resizebox{\linewidth}{!}{%
\begin{tikzpicture}[
    box/.style={draw, rounded corners, align=center, minimum height=9mm,
                inner sep=3pt, font=\footnotesize},
    frozen/.style={box, fill=blue!7},
    trained/.style={box, fill=orange!14},
    op/.style={box, fill=green!8},
    data/.style={box, fill=black!5},
    arr/.style={-{Latex[length=2mm]}, semithick},
    lbl/.style={font=\scriptsize, inner sep=1.5pt}
  ]
  \node[data]    (clip)  at (0,0)       {Driving\\clip};
  \node[frozen]  (enc)   at (2.5,0)     {Frozen\\V-JEPA~2\\encoder};
  \node[trained] (pred)  at (5.5,1.0)   {Predictor\\head};
  \node[op]      (cos)   at (8.3,0.2)   {Cosine\\distance};
  \node[data]    (score) at (11.3,0.2)  {Review-value\\score, rank};
  \node[trained] (probe) at (2.5,-1.6)  {Supervised\\probe\\(diagnostic)};

  \draw[arr] (clip) -- (enc);
  \draw[arr] (enc.east) |- (pred.west) node[lbl, pos=0.72, above] {masked};
  \draw[arr] (pred.east) -| (cos.north) node[lbl, pos=0.28, above] {predicted};
  \draw[arr] (enc.east) -- (cos.west) node[lbl, pos=0.5, below] {target embedding};
  \draw[arr] (cos) -- (score);
  \draw[arr] (enc.south) -- (probe.north);
  \draw[arr, dashed] (probe.east) -| (score.south);
\end{tikzpicture}%
}
\caption{Method overview. A frozen V-JEPA~2 encoder embeds each clip; a trained
predictor reconstructs masked embeddings, and the cosine distance to the target
embedding is the unsupervised review-value score. A supervised probe on the same
frozen embeddings (dashed) serves as a labeled diagnostic. Only the orange
components are trained.}
\label{fig:pipeline}
\end{figure}
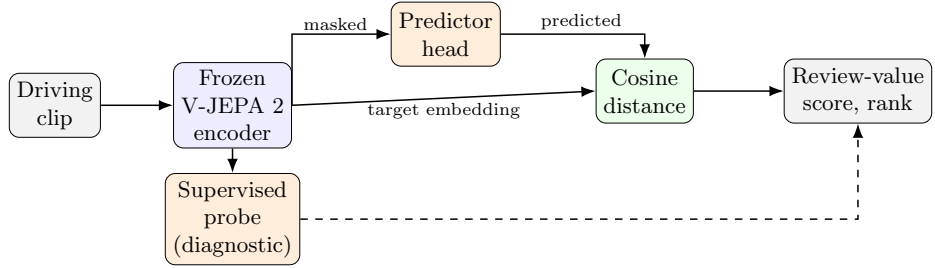

The prior work above motivates a concrete pipeline: take a strong frozen video
representation, learn to predict masked embeddings on in-domain footage, and use
the prediction residual as a triage score. Figure~\ref{fig:pipeline} gives an
overview; this section formalizes the triage problem and then details each
component---the frozen encoder, the masked-embedding predictor, the novelty
score, and the supervised and no-training baselines we compare against.

\subsection{Problem Formulation}
Let $\mathcal{C}=\{c_1,\dots,c_N\}$ be a pool of driving clips. No labels are
available for ranking; each clip nonetheless has a hidden ground-truth binary
review-worthiness label $y(c)\in\{0,1\}$, used only to evaluate the ranking, and
a provenance attribute $d(c)\in\mathcal{D}$ denoting the dataset it originates
from. A triage method is a scoring function
$s:\mathcal{C}\to\mathbb{R}$; ranking clips by decreasing $s$ should place
review-worthy clips first. We measure ranking quality by Average Precision,
\begin{equation}
  \mathrm{AP}(s) \;=\; \frac{1}{P}\sum_{k=1}^{N}
      \mathrm{Precision}@k \,\cdot\, \mathbf{1}\!\left[\,y(c_{(k)})=1\,\right],
\end{equation}
where $c_{(k)}$ is the clip ranked $k$-th by $s$, $P=\sum_{c}y(c)$ is the number
of positives, and $\mathrm{Precision}@k$ is the precision among the top-$k$ clips. The
chance level of AP is approximately the positive rate $P/N$ (exactly so in
expectation as $N\to\infty$; for finite $N$ the expected AP of a random ranking
sits slightly above it).

A frozen encoder $f$ maps a clip to an embedding $z=f(c)\in\mathbb{R}^{1024}$.
Every method we study derives its score $s$ from $f$ and differs only in what is
placed between the embedding and the score: a trained predictor, an untrained
predictor, a density estimate, or a supervised classifier.

\subsection{Data Split}
In practice, a triage system is most valuable on data \emph{unlike} what a fleet
has already curated. A model is trained on a limited, fixed corpus, whereas the
footage that later streams in may originate from different vehicles, cameras,
cities, or weather conditions---often with different sensors and color grading.
A natural way to test whether a score surfaces such material is therefore to
source clearly review-worthy clips from driving datasets \emph{other} than the
training source---here Waymo and BDD100K---and to treat the abundant routine
footage of the training dataset (nuScenes) as the negatives. This protocol is intuitive, mirrors deployment, and is easy to
assemble, so we adopt it as our primary evaluation. 

Concretely, we draw clips from three driving datasets. Let $\mathcal{C}_1$ denote
our \emph{in-domain} dataset, nuScenes, and $\mathcal{C}_2$ the \emph{external}
datasets, Waymo and BDD100K. The predictor head is pre-trained on unlabeled clips
from $\mathcal{C}_1$ only; the external set $\mathcal{C}_2$ is held out from
training entirely and is used purely to supply review-worthy clips at evaluation
time. We fix nuScenes as the sole in-domain source $\mathcal{C}_1$ because our
within-dataset review-worthiness labels exist only for it; letting each of the
three datasets play the role of $\mathcal{C}_1$ in turn would require comparable
labels in Waymo and BDD100K, which we do not have. For the cross-dataset benchmark, the ranked pool $\mathcal{C}$ comprises
routine clips from $\mathcal{C}_1$ and review-worthy clips drawn predominantly
from $\mathcal{C}_2$; for the within-nuScenes benchmark, $\mathcal{C}$ is drawn
from $\mathcal{C}_1$ alone. In short, $\mathcal{C}_1$ and $\mathcal{C}_2$ name the data \emph{sources},
while the predictor's (unlabeled) training set and each evaluation pool
$\mathcal{C}$ are subsets drawn from them.

A useful score should, however, reflect genuine review value rather than mere
differences in appearance between data sources. This is a distinction we will return to when
analyzing the results.

\subsection{Frozen Encoder and Mean-Pooled Clip Embeddings}
We use a frozen V-JEPA~2 video encoder~\cite{vjepa2}
(\texttt{facebook/vjepa2-vitl-fpc64-256}) as a fixed feature extractor; its
weights are never updated. For a clip, the
encoder produces a sequence of $1024$-dimensional token embeddings over
two-frame tubelets. We mean-pool these tubelet embeddings into a single
$1024$-dimensional clip vector, which serves both as the prediction target for the
masking objective and as the input feature for the supervised probe. Freezing
the encoder isolates the question we seek to answer: given a fixed, strong video
representation, can a self-supervised masking objective extract a triage signal?

\subsection{Predictor Head and Masking Transform}
On top of the frozen encoder we attach a small trainable predictor head. Each clip is processed as a sequence of two-frame tubelets. During training, a masking
transform zeroes a fraction (mask ratio $0.5$) of the $16\times16$ spatial
patches within each tubelet. The frozen encoder embeds both the masked tubelet
and the unmasked (clean) tubelet into $1024$-d pooled vectors, and the predictor
maps the masked embedding (together with the scalar masked fraction) to an
estimate of the clean embedding. Only the predictor head is trained; the encoder
remains frozen throughout.

\subsection{JEPA Embedding-Prediction Objective}
The predictor is trained with an L1 joint-embedding prediction loss in the
encoder's representation space. For a tubelet, let $z$ denote the frozen-encoder
embedding of the clean tubelet (the target) and $\hat{z}=g(\tilde{z})$ the
predictor $g$ applied to the embedding $\tilde{z}$ of the masked tubelet. The
objective is the L1 distance between the two embeddings,
\begin{equation}
  \mathcal{L} \;=\; \frac{1}{N}\sum_{i=1}^{N}
      \bigl\lVert \hat{z}_i - z_i \bigr\rVert_1,
\end{equation}
averaged over the $N$ tubelets in a batch. Both $z$ and $\tilde{z}$ are produced
by the frozen encoder, so the loss is a pure embedding-prediction (JEPA-style)
objective with no pixel reconstruction, in contrast to masked autoencoders that
reconstruct in input space~\cite{m3ae}.

\subsection{Review-Value Novelty Score}
At inference, we convert prediction error into a per-clip novelty score. For a
clip we form the predicted and target embeddings and define the
\emph{review-value score} as the cosine distance between them,
\begin{equation}
  \text{review-value score} \;=\; 1 - \cos\bigl(\text{predicted},\,
      \text{target}\bigr),
\end{equation}
so that clips whose masked content is poorly predicted (high cosine distance)
receive high scores. The per-tubelet cosine distances are averaged over a clip,
and ranking clips by this score gives the unsupervised triage ordering we
evaluate.

\subsection{Supervised Probe}
As a diagnostic for whether the triage signal exists in the representation at
all, we train a logistic-regression classifier on the frozen mean-pooled clip
embeddings to predict the review-worthy label. The probe uses the same frozen
$1024$-d features as the novelty score; only its supervision differs. It is not
a proposed deployment method---it requires labels---but it measures how much
review-worthiness information is linearly accessible in the frozen representation.

\subsection{No-Training Baselines}
To check whether predictor training contributes anything beyond the frozen
encoder, we compare against three simple baselines that require no learned predictor:
\begin{itemize}
  \item \textbf{Masked-gap.} Score a clip by
        $1 - \cos\bigl(\text{encoder(masked)},\,\text{encoder(clean)}\bigr)$,
        i.e.\ how much masking perturbs the embedding, with no predictor head.
  \item \textbf{kNN-density.} A leave-one-out cosine $k$-nearest-neighbor
        novelty score: clips far (in cosine distance) from their neighbors in
        embedding space are scored as more novel.
  \item \textbf{Untrained-predictor.} The same predictor head architecture as
        above but with random initialization (no training), used to produce a
        novelty score in the same way as the trained predictor.
\end{itemize}

\section{Experimental Setup}
\label{sec:setup}

With the scoring pipeline and baselines defined, we now specify how they are
trained and evaluated: the driving datasets, the two benchmarks that
operationalize the confounded and fair protocols, the labeling procedure for the
supervised probe, the metric suite, and the compute budget.

\subsection{Data}
Our primary data source is the nuScenes dataset~\cite{nuscenes}, using the front
camera (\texttt{CAM\_FRONT}) over the
\texttt{trainval01}--\texttt{trainval04} splits. We additionally use clips from
the Waymo Open Dataset~\cite{waymo} and BDD100K~\cite{bdd100k} to construct the
cross-dataset benchmark described below. All three are large-scale driving
datasets of egocentric, forward-facing video---nuScenes ($1{,}000$ scenes of
$20$\,s, collected in Boston and Singapore), the Waymo Open
Dataset (${\sim}1{,}150$ segments of $20$\,s across several U.S. cities), and
BDD100K ($100{,}000$ clips of $40$\,s spanning diverse cities, weather, and times
of day)---but they differ in sensor, resolution, and color grading. That
appearance gap is precisely what the cross-dataset protocol risks rewarding.

\subsection{Training Sources: Samples vs.\ Sweeps}
nuScenes provides two frame streams per scene: \texttt{samples}, the annotated
keyframes at roughly $2$\,Hz, and \texttt{sweeps}, the dense intermediate frames
at roughly $12$\,Hz. To test whether the source or the temporal extent of the
pretext data affects triage, we train the predictor under a $2\times2$ design
crossing source (\texttt{samples} vs.\ \texttt{sweeps}) with clip length
($16$ vs.\ $32$ frames), each over $5$ seeds. We additionally train a
\texttt{sweeps-all} condition that pools \texttt{sweeps} from
\texttt{trainval01}--\texttt{trainval04} (roughly $5\times$ the data of a single
split) to probe whether more unlabeled pretext data helps. Results for this
configuration sweep appear in Table~\ref{tab:sweep}.

\subsection{Benchmarks}
The two benchmarks introduced earlier are realized concretely as follows. The
\emph{cross-dataset} benchmark contains $64$ clips: $32$ review-worthy positives
($13$ from Waymo, $9$ from BDD100K, $10$ from nuScenes) and $32$ routine nuScenes
negatives, giving a chance AP of $0.50$. The \emph{within-nuScenes} benchmark is
a held-out set of $83$ nuScenes clips with $20$ positives, giving a chance AP of
$0.24$.

\subsection{Human Labeling}
For the supervised probe we additionally label a training pool of $437$ nuScenes
clips ($381$ newly annotated, plus $56$ pre-existing labels) spanning $38$
scenes, $21\%$ of them positive. Two annotators labeled the clips for
review-worthiness. Scenes are held disjoint between this pool
and the $83$-clip within-nuScenes benchmark, so no scene appears in both and
so that there is no scene-level leakage.

\subsection{Metrics}
We report a suite of ranking and detection metrics. Average Precision (AP) is the
precision averaged over the ranks at which review-worthy clips appear, and PR-AUC
is the area under the precision--recall curve; both summarize how well the ranking
concentrates review-worthy clips near the top. ROC-AUC is the area under the
receiver-operating-characteristic curve, where $0.5$ denotes chance. NDCG
(normalized discounted cumulative gain) rewards placing review-worthy clips early
in the ranking and discounts those buried deeper. Precision@$10$ and Recall@$10$
are the precision and recall within the top ten ranked clips, the regime that
matters to a reviewer who inspects only a short prefix of the queue. Because AP
and its chance level depend on the positive rate, we report the chance \emph{AP
floor} (the positive rate) for each benchmark alongside the measured values: a
triage method is only interesting if it exceeds this floor.

\subsection{Hyperparameter Selection}
Predictor hyperparameters were fixed by a preliminary full-factorial sweep over
mask ratio $\in\{0.5,\,0.75\}$, predictor hidden width $\in\{512,\,1024\}$, and
learning rate $\in\{10^{-4},\,3\times10^{-4}\}$, with three random seeds per cell
($2{\times}2{\times}2{\times}3=24$ runs). We select the configuration with the
highest Average Precision (mask ratio $0.5$, hidden width $512$, learning rate
$10^{-4}$), adopt it unchanged, and do \emph{not} re-tune on the fair benchmark.
Because every configuration we examine sits near the chance floor
(Table~\ref{tab:sweep}), this choice does not affect our conclusions.

\subsection{Training Details}
The V-JEPA~2 encoder is frozen for all experiments. The predictor head is
trained for $25$ epochs with the L1 embedding-prediction loss, repeated over $5$
random seeds. For the supervised probe we use scene-disjoint cross-validation
with $10$ repeats, reporting mean$\pm$std. All splits respect scene boundaries
so that train and benchmark scenes never overlap.

\subsection{Compute Cost}
All runs use a single NVIDIA V100 (32\,GB). Table~\ref{tab:compute} reports the
training time, peak GPU memory, scoring throughput, and estimated scoring energy
for each configuration. Because the encoder is frozen and only the small
predictor head is trained, the cost is modest: peak memory stays near $2$\,GB and
training a configuration takes between half an hour and roughly four hours,
scaling with the size of the pretext set.

\begin{table}[tbp]
\caption{Compute cost on a single NVIDIA V100 (32\,GB). Training time covers $25$
epochs of the predictor (the encoder is frozen); throughput and energy are
measured over a fixed $663$-clip timing pass, with energy estimated at $75$\,W}
\label{tab:compute}
\centering
\scriptsize
\setlength{\tabcolsep}{6pt}
\begin{tabular}{lcccc}
\toprule
Configuration & Train (min) & Peak GPU (MB) & Throughput (clips/s) & Score energy (kJ) \\
\midrule
\texttt{samples\_16} & 31  & 1898 & 4.4 & 11.2 \\
\texttt{samples\_32} & 59  & 1900 & 4.7 & 10.5 \\
\texttt{sweeps\_16}  & 146 & 2057 & 4.5 & 11.2 \\
\texttt{sweeps\_32}  & 250 & 2027 & 5.3 & 9.4  \\
\bottomrule
\end{tabular}
\end{table}

\section{Results}
\label{sec:results}

We first contrast the two benchmarks on a single checkpoint, then trace the
inflated cross-dataset number to the novelty score behaving as a domain detector,
and finally sweep the self-supervised configurations to confirm that none of them
escapes the chance floor within nuScenes.

\begin{figure}[tbp]
\centering
\includegraphics[width=\linewidth]{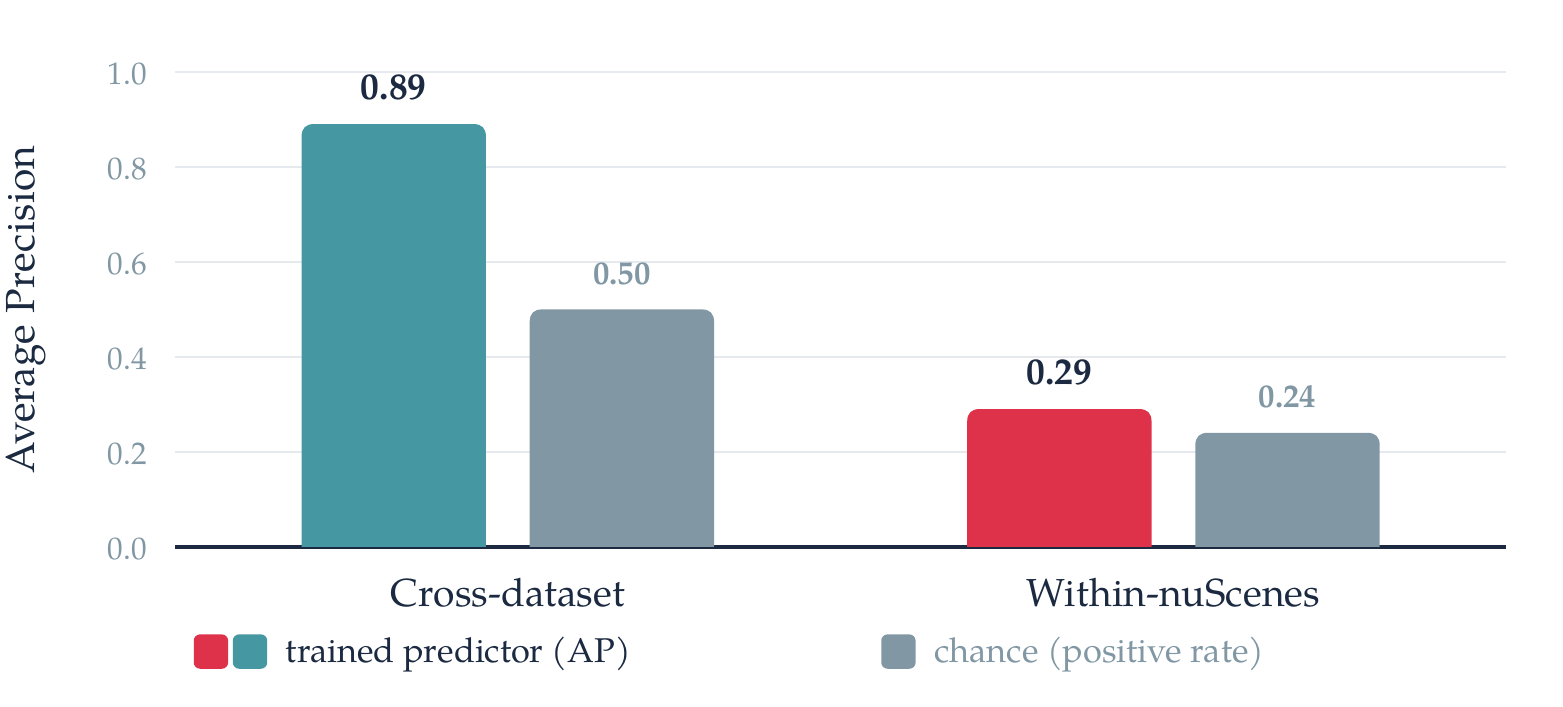}
\caption{Same model, two evaluations. One trained checkpoint scores AP~$0.89$ on
the cross-dataset benchmark (chance~$0.50$) but AP~$0.29$ on the fair
within-nuScenes benchmark (chance~$0.24$).}
\label{fig:two-evals}
\end{figure}

\subsection{Provenance Affects Results}
A single trained predictor checkpoint scores AP~$0.89$ on the cross-dataset
benchmark, well above its $0.50$ chance level (Figure~\ref{fig:two-evals}). On
its own this would read as a strong unsupervised triage result. The same
checkpoint scores AP~$0.29$ on the fair within-nuScenes benchmark, against a
chance floor of $0.24$, which is no better than ranking the clips at random. The
decisive difference between the two evaluations is whether dataset provenance
is confounded with the label.

\subsection{Novelty Is a Domain Detector}
\begin{figure}[tbp]
\centering
\includegraphics[width=\linewidth]{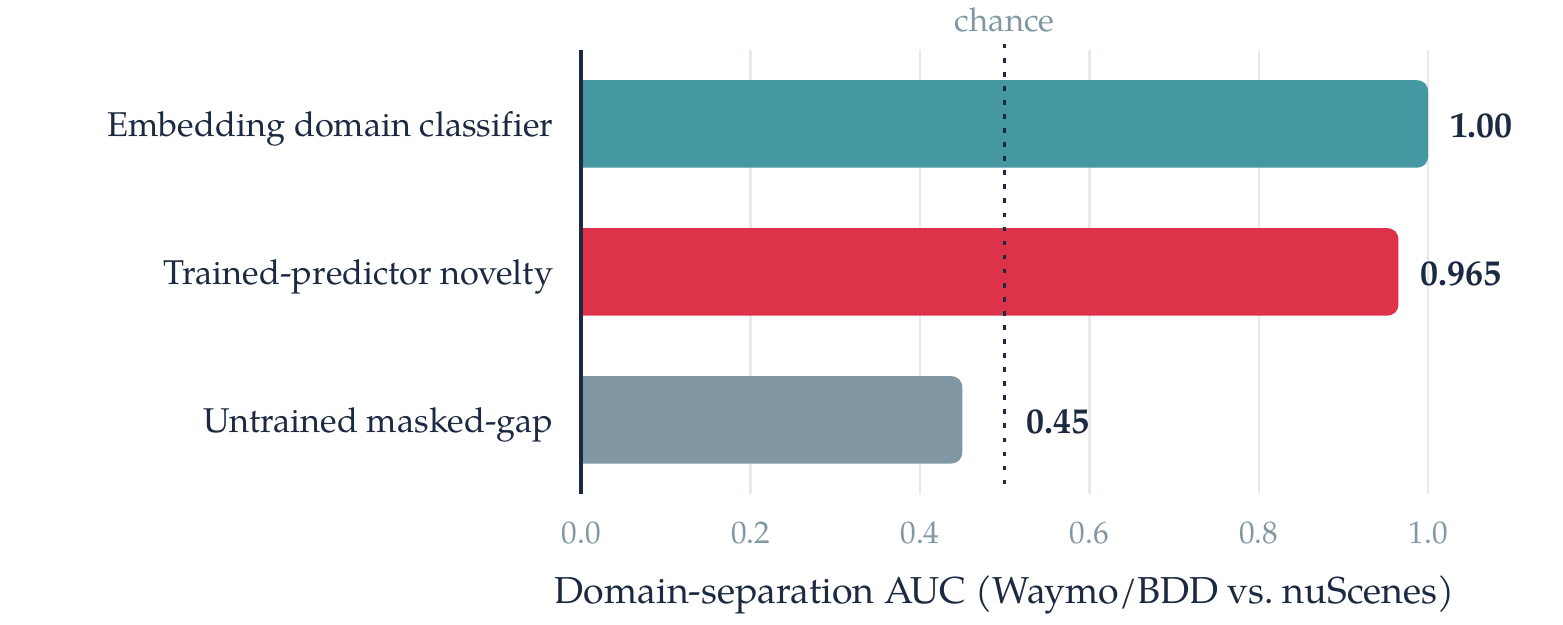}
\caption{Novelty is a domain detector. On the frozen embeddings, a logistic
provenance classifier (AUC~$1.00$) and the trained-predictor novelty score
(AUC~$0.965$) both separate Waymo/BDD from nuScenes, while the untrained
masked-gap score is near chance (AUC~$0.45$; dashed line, AUC$=0.5$).}
\label{fig:domain-detector}
\end{figure}

The reason is that the novelty score has learned to recognize the dataset
(Figure~\ref{fig:domain-detector}). A logistic classifier on the frozen
embeddings separates Waymo/BDD from nuScenes with AUC~$1.00$ ($5$-fold
cross-validation), and the
trained-predictor novelty score tracks that same split almost as well
(AUC~$0.965$). The untrained masked-gap score does not: its domain AUC is
$0.45$, near chance. The per-clip averages say the same thing, with mean
trained-predictor novelty of $0.027$ on nuScenes against $0.103$ on Waymo/BDD.
Training the predictor on nuScenes teaches it to assign higher novelty to
out-of-domain footage, which is exactly the cue the cross-dataset benchmark
rewards.

This explains the cross-dataset result. Its positives come mostly from the
external datasets and its negatives entirely from nuScenes, so the
review-worthiness label is largely a function of provenance. In the idealized
case where every positive is external,
\begin{equation}
  y(c) \;=\; \mathbf{1}\!\left[\,d(c)\in\{\text{Waymo},\,\text{BDD}\}\,\right],
  \label{eq:confound}
\end{equation}
any score that separates the domains, meaning one with high AUC for predicting
$d$, also achieves high $\mathrm{AP}(s)$ for $y$, whether or not it captures
within-domain review value. That is what we see. The cross-dataset AP of $0.89$
measures domain shift rather than review value, and it falls short of a perfect
$1.0$ only because $10$ of the $32$ positives are themselves nuScenes clips,
which a pure domain detector cannot place above the nuScenes negatives.

\subsection{Fair-Benchmark Triage}
\begin{figure}[tbp]
\centering
\includegraphics[width=\linewidth]{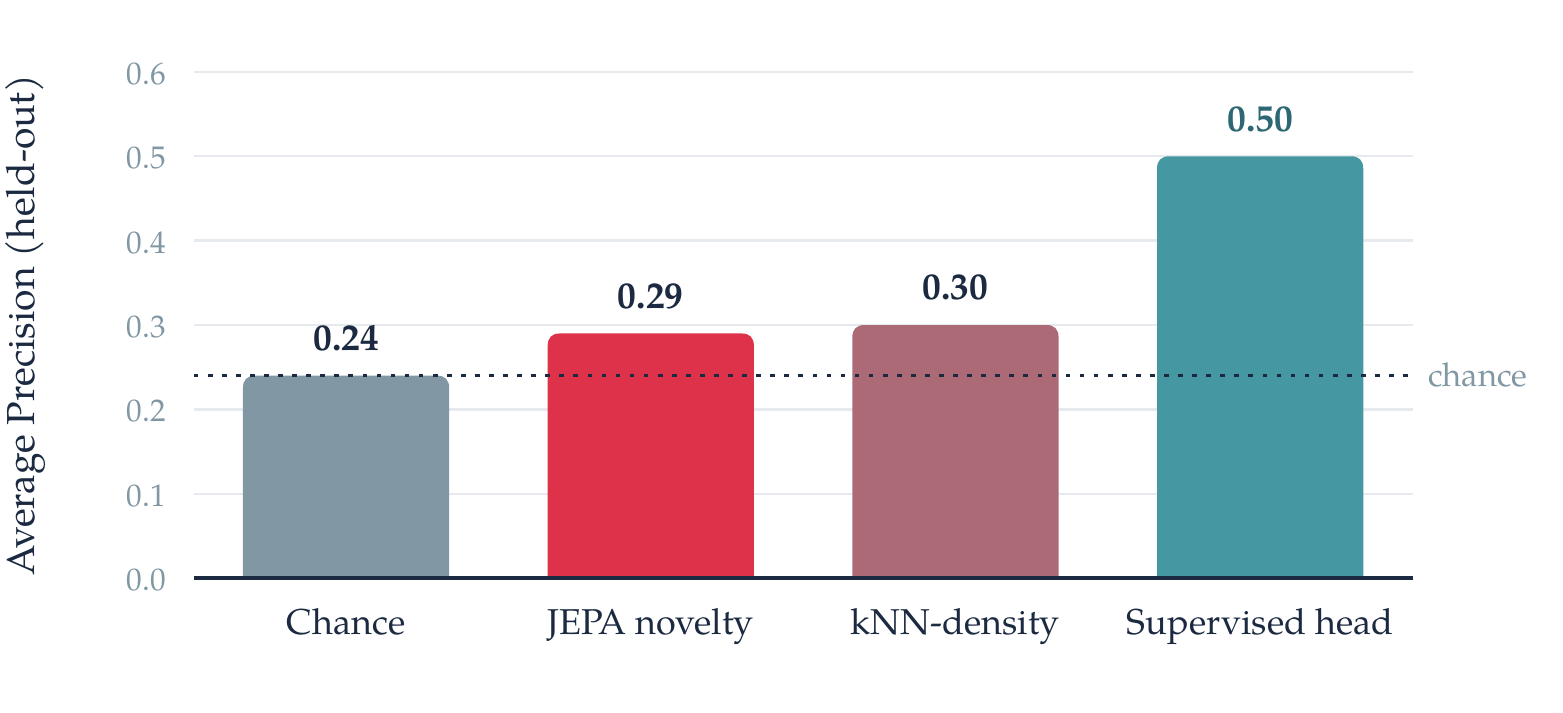}
\caption{Fair-benchmark triage (held-out 83-clip set). Unsupervised JEPA novelty
(AP~$0.29$) and kNN-density (AP~$0.30$) barely exceed the chance floor
(AP~$0.24$, dashed line); a supervised probe on the same embeddings reaches
AP~$0.50$.}
\label{fig:fair-triage}
\end{figure}

On the held-out $83$-clip benchmark (chance AP~$0.24$), unsupervised JEPA
novelty reaches AP~$0.29$ and kNN-density AP~$0.30$, both barely above the floor
(Figure~\ref{fig:fair-triage}). A supervised logistic probe on the same frozen
embeddings reaches AP~$0.50$, roughly twice chance. The signal a reviewer cares
about is present in the representation; the unsupervised masking objective simply
does not surface it.

\subsection{Main Results Table}
Table~\ref{tab:main} reports the full ranking-metric suite for both the training
pool (scene-disjoint cross-validation) and the held-out benchmark. Across every
metric the unsupervised novelty and kNN-density scores stay close to chance,
while the supervised head is the only method that moves substantially off the
floor, so the gap is not an artifact of any single measure.

\begin{table}[tbp]
\caption{Triage results across the full ranking-metric suite, for the training
pool (scene-disjoint cross-validation, $10$ repeats) and the held-out $83$-clip
benchmark. Chance AP is the positive rate ($0.21$ pool, $0.24$ held-out); the
best value in each column and panel is in bold. Pool supervised-head values are
the mean over $10$ scene-disjoint CV repeats (AP~$0.411\pm0.028$, ROC-AUC
$0.662\pm0.026$); the unsupervised novelty and kNN-density scores are
deterministic. The supervised head is a labeled diagnostic, not a deployable
unsupervised method}
\label{tab:main}
\centering
\resizebox{\linewidth}{!}{%
\setlength{\tabcolsep}{17pt}%
\begin{tabular}{lcccccc}
\toprule
Method & AP & PR-AUC & ROC-AUC & NDCG & Precision@10 & Recall@10 \\
\midrule
\multicolumn{7}{l}{\textit{(a) Training pool (scene-disjoint cross-validation)}}\\
\midrule
Chance (positive rate)     & 0.213 & --    & 0.500 & --    & 0.213 & 0.023 \\
Unsupervised JEPA novelty  & 0.315 & 0.310 & 0.624 & 0.751 & 0.200 & 0.022 \\
kNN-density                & 0.327 & 0.322 & 0.636 & 0.755 & 0.200 & 0.022 \\
Supervised head            & \textbf{0.411} & \textbf{0.406} & \textbf{0.662} & \textbf{0.819} & \textbf{0.700} & \textbf{0.075} \\
\midrule
\multicolumn{7}{l}{\textit{(b) Held-out 83-clip benchmark}}\\
\midrule
Chance (positive rate)     & 0.241 & --    & 0.500 & --    & 0.241 & 0.120 \\
Unsupervised JEPA novelty  & 0.288 & 0.261 & 0.533 & 0.643 & 0.300 & 0.150 \\
kNN-density                & 0.302 & 0.285 & 0.629 & 0.604 & 0.200 & 0.100 \\
Supervised head            & \textbf{0.499} & \textbf{0.489} & \textbf{0.663} & \textbf{0.831} & \textbf{0.600} & \textbf{0.300} \\
\bottomrule
\end{tabular}}
\end{table}

\subsection{Self-Supervised Configuration Sweep}
Table~\ref{tab:sweep} varies the self-supervised configuration on the fair
benchmark (chance AP~$0.24$). Changing the sampling scheme (\texttt{samples}
vs.\ \texttt{sweeps}) and the clip length ($16$ vs.\ $32$ frames) leaves the AP
flat and clustered just above chance, and training on $5\times$ more data
(\texttt{sweeps-all}) reaches only AP~$0.28$. The no-training baselines, with AP
of $0.277$ (masked-gap), $0.301$ (untrained predictor), and $0.304$
(kNN-density), do no worse than any trained variant. Training the predictor
therefore adds essentially nothing to within-domain triage.

\begin{table}[t]
\caption{Self-supervised configuration sweep on the fair benchmark. Chance
AP~$=0.24$. Trained-predictor variants (top) and no-training baselines (bottom)
all remain near chance, and no configuration consistently outperforms the
no-training baselines}
\label{tab:sweep}
\centering
\footnotesize
\begin{tabular}{lc}
\toprule
Configuration & AP \\
\midrule
\multicolumn{2}{l}{\textit{Self-supervised (trained predictor)}}\\
\texttt{samples\_16}                 & $0.273 \pm 0.006$ \\
\texttt{samples\_32}                 & $0.269 \pm 0.003$ \\
\texttt{sweeps\_16}                  & $0.263 \pm 0.007$ \\
\texttt{sweeps\_32}                  & $0.256 \pm 0.008$ \\
\texttt{sweeps-all} ($5\times$ data) & $0.280$ \\
\midrule
\multicolumn{2}{l}{\textit{No-training baselines}}\\
Masked-gap          & $0.277$ \\
Untrained-predictor & $0.301$ \\
kNN-density         & $0.304$ \\
\bottomrule
\end{tabular}
\end{table}

\section{Discussion}
\label{sec:discussion}

\begin{figure}[t]
\centering
\includegraphics[width=\linewidth]{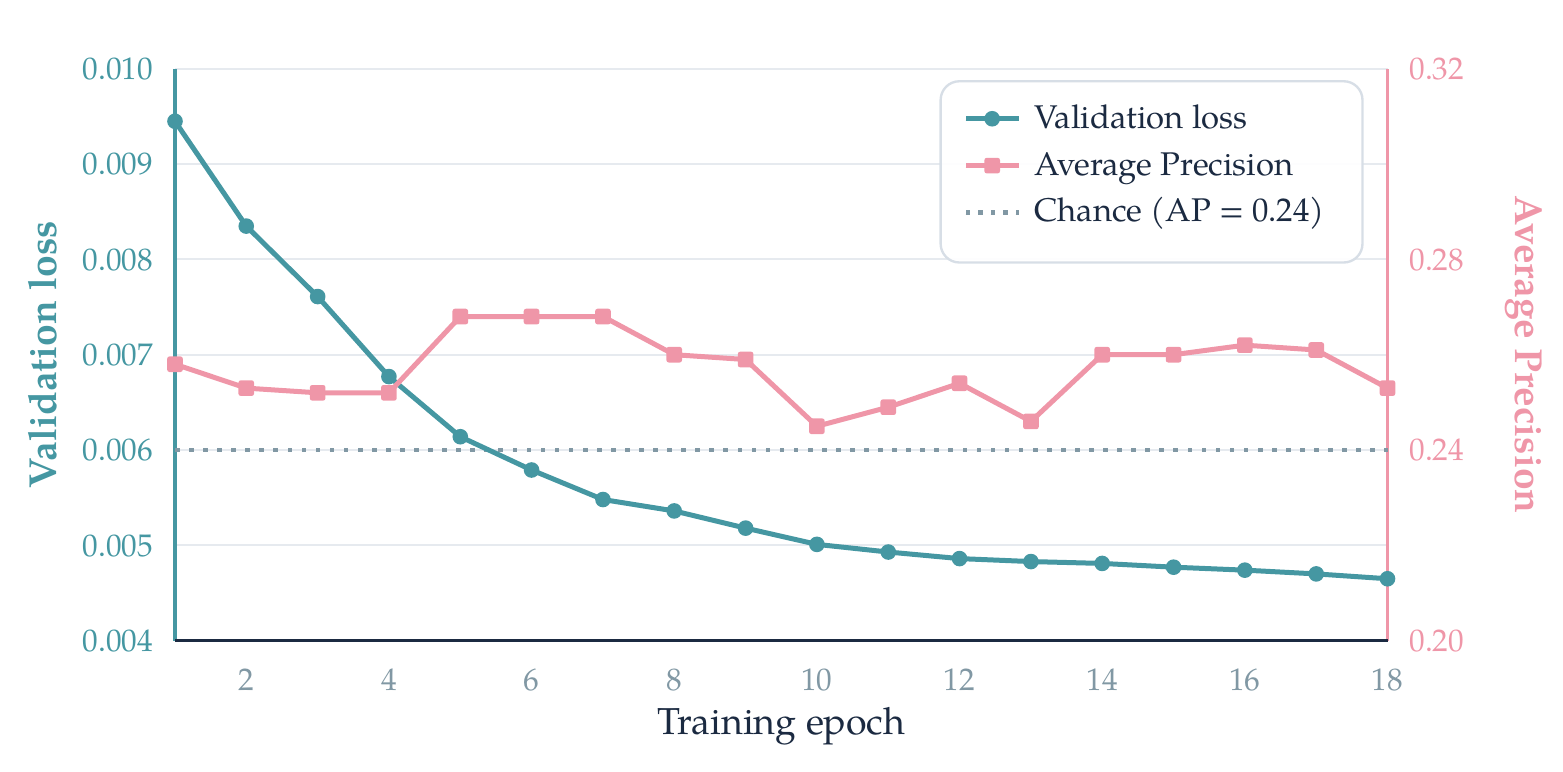}
\caption{Training does not help triage. For one representative run, the
masking-objective validation loss falls and plateaus (blue, left axis) while
held-out AP stays near the chance floor (red, right axis; dashed at AP$=0.24$),
which makes undertraining an unlikely explanation. Reported models train for
$25$ epochs; the $18$ epochs shown already reach the loss plateau.}
\label{fig:convergence}
\end{figure}

The usual levers for improving a weak pretext model, namely a better-tuned
masking loss, denser temporal sampling, and more pretext data, did not help
within-domain triage; every configuration stayed near the chance floor.
Undertraining is not the explanation either: over the course of training the
validation loss falls and then plateaus while the held-out AP never leaves the
chance floor (Figure~\ref{fig:convergence}). The trained predictor is better
described as a domain detector, and the reason is visible in the embeddings
themselves. We hypothesize that provenance is a highly linearly separable axis
of variation in the frozen feature space---a linear classifier separates the
source datasets almost perfectly (AUC~$1.00$, Figure~\ref{fig:domain-detector})---and a masking
objective that minimizes average embedding-prediction error is drawn toward
exactly this kind of high-variance, globally consistent structure. Predicting the
dataset-characteristic component of an embedding lowers the loss far more reliably
than predicting the rare, spatially or temporally localized cues that make an
individual clip worth reviewing, so the prediction residual ends up measuring how
far a clip sits from the in-domain manifold rather than how unusual its content is
within that manifold. Training then succeeds only when review-worthiness happens
to be confounded with provenance.

The failure is therefore one of extraction rather than information. A supervised
linear probe on the same frozen embeddings roughly doubles the chance AP, so the
signal a reviewer wants is present in the representation; the masking objective
simply does not learn to predict it. Because the encoder is frozen throughout,
this is a statement about the objective acting on a fixed representation: freezing
is what lets us attribute the gap to the predictor rather than to a shifting
backbone, but it also leaves open whether end-to-end fine-tuning would reshape the
feature space enough to decouple novelty from provenance in the first place.

The consequence extends beyond this particular encoder. Whenever positives and
negatives are drawn from different datasets, a score with even mild sensitivity to
provenance can post strong headline numbers while carrying no within-domain triage
value, and the gap stays invisible unless a same-source benchmark is run
alongside. The remedy is cheap: decorrelate review value from provenance by
sampling both positives and negatives from every source, and report a supervised
probe on the frozen features as a ceiling that distinguishes a genuinely absent
signal from one the objective merely fails to surface. As self-supervised video
foundation models are increasingly reused off the shelf for autonomous-driving
data curation, we expect this confound to recur, and a same-source control to
remain the cheapest way to catch it.

\section{Limitations and Future Work}
\label{sec:limitations}

Our fair benchmark is small, an $83$-clip held-out set with $20$ positives, so
the within-domain estimates carry meaningful variance. The labels come from two
annotators, which still leaves room for label noise and subjective bias. A
better evaluation would use a balanced multi-dataset benchmark with both positive
and negative clips drawn from each dataset, so that review value and provenance
are decorrelated by construction; assembling one requires per-dataset
review-worthiness labels we do not yet have. A further limitation is on the
representation side: we mean-pool the tubelet embeddings into a single clip
vector, which dilutes exactly the spatially and temporally localized cues that
often make an event review-worthy, so a pooled score may wash out short or
small-region anomalies.

The most important limitation is one of \emph{scope}. Our evidence comes from a
single self-supervised configuration---a frozen V-JEPA~2 encoder, one predictor
head, one masking strategy, and one evaluation protocol---so our conclusions
apply to this instantiation rather than to masking-based or self-supervised
representation learning as a whole. Three axes of generalization are, in our
view, the priority for future work. First, \emph{other encoders}: the same
diagnostic should be run on widely used self-supervised video and image--text
representations---VideoMAE, DINOv2, VICReg, MoCo-v3, SimCLR, BYOL, and CLIP-based
video embeddings---to establish whether the provenance confound is specific to
JEPA-style prediction error or a broader property of frozen self-supervised
features. Second, \emph{adaptation}: because the encoder is frozen throughout,
we cannot separate the failure of the objective from the decision to fix the
representation; fine-tuning the encoder end-to-end, and replacing spatial masking
with a temporal or future-frame prediction objective, are the natural
counterfactuals. Third, \emph{other downstream tasks}: review-worthiness ranking
is one instance of novelty-driven curation, and validating the observation on
anomaly detection, accident anticipation, rare-event retrieval, safety-critical
event identification, and corner-case mining would test how far the diagnostic
generalizes. On the analysis side, feature-space visualizations (t-SNE/UMAP),
embedding-distribution and attention-map inspection, and qualitative case studies
of the highest- and lowest-scored clips would further explain \emph{why} the
objective learns dataset-specific structure. Retaining localized (un-pooled)
structure in the score, rather than the mean-pooled clip vector noted above, is a
complementary direction on the representation side.

\section{Conclusion}
\label{sec:conclusion}

Self-supervised video foundation models are increasingly put forward as a
label-free way to curate the enormous volume of footage that autonomous-driving
fleets collect, and prediction-error ``novelty'' is one of the most natural
signals to build such triage on. We set out to do exactly this with an
unsupervised JEPA-novelty score on a frozen V-JEPA~2 encoder. Across datasets it
looked strong, but the same checkpoint fell to chance on a fair within-nuScenes
benchmark, because the score had learned to detect the dataset rather than the
review value; a supervised probe on the identical frozen embeddings recovered a
clear signal, so the information was there all along. The failure we document is
therefore one of extraction, not of representation, and---for this
instantiation---one of evaluation as much as of method: a cross-dataset protocol
can badly overstate unsupervised triage, and a strong representation does not
guarantee that a given self-supervised objective will surface the property one
actually cares about.

We deliberately frame this as a diagnostic rather than a negative verdict on
self-supervised curation. As the field moves toward ever-larger frozen video
encoders and reuses their features off the shelf, the cheap controls this study
argues for---drawing positives and negatives from every source so that
provenance and review value are decorrelated, and reporting a supervised probe as
an extraction ceiling---become correspondingly more valuable. Whether the same
confound holds for other encoders, for fine-tuned backbones, and for downstream
tasks beyond review-worthiness ranking is the open question our findings raise,
and the one we hope to answer next.

\section*{Declarations}

\paragraph{Use of generative AI.} A large language model was used to help edit
and refine wording and to assist in generating the figures and tables. The
authors designed the study, ran all experiments, and take full responsibility for
the content of this paper.

\paragraph{Code availability.} The code for the experiments, figures, and tables
is available at \url{https://github.com/shamikkarkhanis/AV-SSL-Optimization-JEPA}.

\bibliographystyle{splncs04}
\bibliography{references}

@article{vjepa2,
  title        = {{V-JEPA 2}: Self-Supervised Video Models Enable Understanding,
                  Prediction and Planning},
  author       = {Assran, Mahmoud and others},
  journal      = {arXiv preprint arXiv:2506.09985},
  year         = {2025},
  eprint       = {2506.09985},
  archivePrefix= {arXiv},
  primaryClass = {cs.CV}
}

@article{m3ae,
  title        = {Multimodal Masked Autoencoders Learn Transferable
                  Representations},
  author       = {Geng, Xinyang and Liu, Hao and Lee, Lisa and Schuurmans, Dale
                  and Levine, Sergey and Abbeel, Pieter},
  journal      = {arXiv preprint arXiv:2205.14204},
  year         = {2022},
  eprint       = {2205.14204},
  archivePrefix= {arXiv},
  primaryClass = {cs.CV}
}

@inproceedings{ijepa,
  title        = {Self-Supervised Learning from Images with a
                  Joint-Embedding Predictive Architecture},
  author       = {Assran, Mahmoud and Duval, Quentin and Misra, Ishan and
                  Bojanowski, Piotr and Vincent, Pascal and Rabbat, Michael and
                  others},
  booktitle    = {Proceedings of the IEEE/CVF Conference on Computer Vision
                  and Pattern Recognition (CVPR)},
  pages        = {15619--15629},
  year         = {2023}
}

@inproceedings{maskfeat,
  title        = {Masked Feature Prediction for Self-Supervised Visual
                  Pre-Training},
  author       = {Wei, Chen and Fan, Haoqi and Xie, Saining and Wu, Chao-Yuan
                  and Yuille, Alan and Feichtenhofer, Christoph},
  booktitle    = {Proceedings of the IEEE/CVF Conference on Computer Vision
                  and Pattern Recognition (CVPR)},
  pages        = {14668--14678},
  year         = {2022}
}

@inproceedings{timesformer,
  title        = {Is Space-Time Attention All You Need for Video Understanding?},
  author       = {Bertasius, Gedas and Wang, Heng and Torresani, Lorenzo},
  booktitle    = {Proceedings of the International Conference on Machine
                  Learning (ICML)},
  year         = {2021}
}

@article{vjepa,
  title        = {Revisiting Feature Prediction for Learning Visual
                  Representations from Video},
  author       = {Bardes, Adrien and Garrido, Quentin and Ponce, Jean and
                  Chen, Xinlei and Rabbat, Michael and LeCun, Yann and
                  others},
  journal      = {arXiv preprint arXiv:2404.08471},
  year         = {2024},
  eprint       = {2404.08471},
  archivePrefix= {arXiv},
  primaryClass = {cs.CV}
}

@inproceedings{nuscenes,
  title        = {{nuScenes}: A Multimodal Dataset for Autonomous Driving},
  author       = {Caesar, Holger and Bankiti, Varun and Lang, Alex H. and
                  Vora, Sourabh and Liong, Venice Erin and Xu, Qiang and
                  others},
  booktitle    = {Proceedings of the IEEE/CVF Conference on Computer Vision
                  and Pattern Recognition (CVPR)},
  pages        = {11621--11631},
  year         = {2020}
}

@inproceedings{waymo,
  title        = {Scalability in Perception for Autonomous Driving:
                  {Waymo} Open Dataset},
  author       = {Sun, Pei and Kretzschmar, Henrik and Dotiwalla, Xerxes and
                  Chouard, Aurelien and Patnaik, Vijaysai and Tsui, Paul and
                  others},
  booktitle    = {Proceedings of the IEEE/CVF Conference on Computer Vision
                  and Pattern Recognition (CVPR)},
  pages        = {2446--2454},
  year         = {2020},
  note         = {arXiv:1912.04838}
}

@inproceedings{bdd100k,
  title        = {{BDD100K}: A Diverse Driving Dataset for Heterogeneous
                  Multitask Learning},
  author       = {Yu, Fisher and Chen, Haofeng and Wang, Xin and Xian, Wenqi
                  and Chen, Yingying and Liu, Fangchen and others},
  booktitle    = {Proceedings of the IEEE/CVF Conference on Computer Vision
                  and Pattern Recognition (CVPR)},
  pages        = {2636--2645},
  year         = {2020}
}

@inproceedings{mae,
  title        = {Masked Autoencoders Are Scalable Vision Learners},
  author       = {He, Kaiming and Chen, Xinlei and Xie, Saining and Li, Yanghao
                  and Doll{\'a}r, Piotr and Girshick, Ross},
  booktitle    = {Proceedings of the IEEE/CVF Conference on Computer Vision
                  and Pattern Recognition (CVPR)},
  pages        = {16000--16009},
  year         = {2022}
}

@inproceedings{videomae,
  title        = {{VideoMAE}: Masked Autoencoders are Data-Efficient Learners
                  for Self-Supervised Video Pre-Training},
  author       = {Tong, Zhan and Song, Yibing and Wang, Jue and Wang, Limin},
  booktitle    = {Advances in Neural Information Processing Systems (NeurIPS)},
  volume       = {35},
  pages        = {10078--10093},
  year         = {2022}
}

@inproceedings{contrastive_video_ssl,
  title        = {Spatiotemporal Contrastive Video Representation Learning},
  author       = {Qian, Rui and Meng, Tianjian and Gong, Boqing and
                  Yang, Ming-Hsuan and Wang, Huisheng and Belongie, Serge and
                  others},
  booktitle    = {Proceedings of the IEEE/CVF Conference on Computer Vision
                  and Pattern Recognition (CVPR)},
  pages        = {6964--6974},
  year         = {2021}
}

@inproceedings{sultani_anomaly,
  title        = {Real-World Anomaly Detection in Surveillance Videos},
  author       = {Sultani, Waqas and Chen, Chen and Shah, Mubarak},
  booktitle    = {Proceedings of the IEEE/CVF Conference on Computer Vision
                  and Pattern Recognition (CVPR)},
  pages        = {6479--6488},
  year         = {2018}
}

@article{bogdoll_ad_survey,
  title        = {Anomaly Detection in Autonomous Driving: A Survey},
  author       = {Bogdoll, Daniel and Nitsche, Maximilian and Z{\"o}llner, J. Marius},
  journal      = {arXiv preprint arXiv:2204.07974},
  year         = {2022},
  eprint       = {2204.07974},
  archivePrefix= {arXiv},
  primaryClass = {cs.CV}
}

@inproceedings{dota,
  title        = {When, Where, and What? A New Dataset for Anomaly Detection
                  in Driving Videos},
  author       = {Yao, Yu and Wang, Xizi and Xu, Mingze and Pu, Zelin and
                  Atkins, Ella and Crandall, David},
  booktitle    = {Proceedings of the IEEE/CVF Winter Conference on Applications
                  of Computer Vision (WACV)},
  year         = {2021},
  note         = {arXiv:2004.03044}
}

@article{ssl_tad,
  title        = {Simplifying Traffic Anomaly Detection with Video
                  Foundation Models},
  author       = {Orlova, Svetlana and Kerssies, Tommie and
                  Englert, Brun{\'o} B. and Dubbelman, Gijs},
  journal      = {arXiv preprint arXiv:2507.09338},
  year         = {2025},
  eprint       = {2507.09338},
  archivePrefix= {arXiv},
  primaryClass = {cs.CV}
}

@inproceedings{tao_vad,
  title        = {Track Any Anomalous Object: A Granular Video Anomaly
                  Detection Pipeline},
  author       = {Huang, Yuzhi and Li, Chenxin and Zhang, Haitao and Lin, Zixu
                  and Lin, Yunlong and Liu, Hengyu and others},
  booktitle    = {Proceedings of the IEEE/CVF Conference on Computer Vision
                  and Pattern Recognition (CVPR)},
  pages        = {8689--8699},
  year         = {2025}
}

\end{document}